\pdfoutput=1

\documentclass[11pt]{article}

\usepackage{emnlp2021}

\usepackage{times}
\usepackage{latexsym}

\usepackage[T1]{fontenc}

\usepackage[utf8]{inputenc}

\usepackage{microtype}

\usepackage{float}
\usepackage{graphicx}
\usepackage{enumitem}

\title{Multi-granular Legal Topic Classification on Greek Legislation}

\author{Christos Papaloukas$^{\;\;\dagger}$\qquad Ilias Chalkidis$^{\ddagger}$ \\ \textbf{Konstantinos Athinaios}$^{\;\dagger}$ \qquad \textbf{Despina-Athanasia Pantazi$^{\;\dagger}$} \qquad
\textbf{Manolis Koubarakis$^{\;\dagger}$} \\
$^{\dagger\;}$National and Kapodistrian University of Athens, Greece  \\
\texttt{[christospap, dpantazi, koubarak]@di.uoa.gr} \\
$^{\ddagger\;}$University of Copenhagen, Denmark \\
\texttt{ilias.chalkidis@di.ku.dk}
\\
}

\begin{document}
\maketitle
\begin{abstract}
In this work, we study the task of classifying legal texts written in the Greek language. We introduce and make publicly available a novel dataset based on Greek legislation, consisting of more than 47 thousand official, categorized Greek legislation resources. We experiment with this dataset and evaluate a battery of advanced methods and classifiers, ranging from traditional machine learning and RNN-based methods to state-of-the-art Transformer-based methods. 
We show that recurrent architectures with domain-specific word embeddings offer improved overall performance while being competitive even to transformer-based models. Finally, we show that cutting-edge multilingual and monolingual transformer-based models brawl on the top of the classifiers’ ranking, making us question the necessity of training monolingual transfer learning models as a rule of thumb. To the best of our knowledge, this is the first time the task of Greek legal text classification is considered in an open research project, while also Greek is a language with very limited NLP resources in general. 
\end{abstract}

\section{Introduction}\label{sec:intro}

In recent years, there has been intensified activity in the adaptation of Artificial Intelligence technologies to the legal domain \cite{Chalkidis2018DeepLI, zhong-etal-2020-nlp, chalkidis-etal-2021-lexglue}, in which legal practitioners are required to analyze and review an overwhelming amount of legal data, mostly being plain text documents. This process requires dedication and an extraordinary level of resources, both concerning human resources along with the use of automated techniques to sift rationally through data. However, more sophisticated automated techniques are able to assist legal experts in making obsolete many labour-intensive manual tasks. These techniques are mostly contained in the area of machine learning and natural language processing (NLP).

With legal text processing being an emerging subarea of NLP, many relevant applications have been derived such as legal topic classification~\cite{Nallapati2008LegalDC,chalkidis-etal-2020-empirical} legal information extraction ~\cite{ONeill2017,Chalkidis2018ObligationAP}, legal entity recognition ~\cite{Chalkidis2017ExtractingCE,Angelidis2018NamedER,leitner2019fine}, court opinion generation and analysis ~\cite{Wang2012HistoricalAO,Ye2018InterpretableCP}, legal judgement prediction ~\cite{Aletras2016PredictingJD, xiao_cail2018_2018,Chalkidis2019NeuralLJ} and many more. However, current legal NLP studies are mainly focused in English and Chinese, with very limited resources being available in other languages.\vspace{-1mm}

Our work focuses on the task of \emph{multi-class legal topic classification}, where the goal is to identify the relevant thematic topic that represents a document. In our case, thematic topics (categories) are available in a multi-level hierarchy from broader to more specialized ones.
The main contributions of our work are listed below:
\vspace{-2mm}
\begin{itemize}[leftmargin=8pt, noitemsep]
\item We introduce Greek Legal Code (\textit{GLC}), a dataset consisting of approx. 47k legal resources from Greek legislation. The origin of \textit{GLC} is ``Permanent Greek Legislation Code - Raptarchis'', a collection of Greek legislative documents classified into multi-level (from broader to more specialized) categories.

\item We study the task of multi-class legal topic classification for Greek legislation by examining a battery of advanced methods, ranging from traditional machine learning techniques and RNN-based methods, to state-of-the-art Transformer-based methods. We discuss the results and lay the groundwork for further research.

\end{itemize}
\vspace{-2mm}

Considering that Greek is a language with few NLP resources, we anticipate that our study will be a significant contribution for the Greek NLP community. To enhance the available NLP resources and foster reproducible results, we make both our code and dataset publicly available.

\section{Related Work}\label{sec:relw}

\citet{Mencia2007} introduced a legal topic classification task using a dataset obtained from the EUR-LEX\footnote{See \href{https://eur-lex.europa.eu/}{https://eur-lex.europa.eu/}.} database, which includes EU laws that have been tagged with EUROVOC concepts. They used multiple binary Perceptrons, one for each label, and multi-label pairwise Perceptrons on top of Bag-of-Words (BoW) representations. While the methods seem primal and inefficient by today’s standards, the EURLEX dataset is widely adopted as a notable benchmark in Large-scale Multi-label Text Classification (LMTC) literature.

~\citet{Nallapati2008LegalDC} were also among the first who investigated the task of text classification in the legal domain where machine learning classifiers such as SVMs were insufficient. They experimented with a dataset of 5.5k US docket entries of court cases. The authors stress the importance of feature selection in such specialized domains and expose the limitations of classifiers relying on BoW featues to capture the intricacies of natural language, widespread in specialized domains such as the legal one.

~\citet{Undavia2018ACS} applied neural networks on legal document classification in a similar task, classification of legal court opinions. They used a dataset of 8k US Supreme Court (SCOTUS) opinions, where they targeted two sub-tasks, depending on the total output categories: 15 broad and 279 finer-grained categories. They experimented with shallow neural networks using different word embeddings, where their best model (word2vec + CNN) scored 72.4\% accuracy in the 15-classes task and 31.9\% accuracy in the 279-classes task. They concluded claiming that an RNN-based network together with domain-specific word embeddings could possibly tackle the task with higher accuracy.

~\citet{Chalkidis2019LargeScaleMT} experimented with several classifiers on a novel dataset of 57k legislative documents (EURLEX57k) from EUR-LEX in English. They demonstrated that that BiGRUs with self attention outperform CNN-based methods that employ the label-wise attention mechanism. Using domain-specific word embeddings and context-sensitive ELMO \cite{peters-etal-2018-deep} embeddings improves the overall performance. Furthermore, the authors experimented with BERT \cite{Devlin2019BERTPO} obtaining the best results 

In a more recent and extended version of this study, ~\citet{Chalkidis2020AnES} evaluated a battery of LMTC methods ranging from RNN-based Label-Wise Attention Networks (LWANs) to Probabilistic Label Trees (PLTs)~\cite{Prabhu2018,Khandagale2019, You2019} and Transformer-based models (BERT, ROBERTA)~\cite{Devlin2019BERTPO,Liu2019RoBERTaAR} on three English datasets: EURLEX57k~\cite{Chalkidis2019LargeScaleMT}, MIMIC-III~\cite{johnson2016mimic} and AMAZON13k~\cite{10.5555/1005332.1005345}. The experimental results show that PLT-based methods outperform LWANs, while Transformer-based approaches surpass state-of-the-art in two out of three datasets. Furthermore, a new state-of-the-art method is introduced which combines BERT and LWAN, giving the best results overall. Furthermore, the case of few and zero-shot learning is studied with new models that leverage the label hierarchy and yield better results.

Following literature, we examine traditional machine learning methods and the RNN-based methods used in ~\citet{Chalkidis2019LargeScaleMT} and ~\citet{Chalkidis2020AnES} and investigate whether these methods can perform equally well in multi-class text classification as they do in the multi-label setting. Furthermore, we examine several BERT-based methods, including multi-lingual models~\cite{Devlin2019BERTPO, Conneau2020UnsupervisedCR} that have not been studied to date in the context of legal NLP.

Most of the preceding efforts focus on the English language and limited studies and resources exist considering languages other than English and Chinese. While recently there is a new wave of studies on NLP tasks focused on the Greek language ~\cite{Athanasiou2017ANG,Papantoniou2020NLPFT,Pitenis2020OffensiveLI,Koutsikakis2020GREEKBERTTG}, there is limited work for legal NLP in Greek~\cite{Angelidis2018NamedER}. To the best of our knowledge, this is the first study on the task of Greek legal text classification, where experiments range from traditional machine learning to transfer learning models.

\section{GLC Dataset}\label{sec:rap47k}

\subsection{Original Data}

The ``Permanent Greek Legislation Code - Raptarchis\footnote{Mr. P. Raptarchis is the original curator of this collection.}'' is a thorough catalogue of Greek legislation since the creation of the Greek state in 1834 until 2015. It includes Laws, Royal and Presidential Decrees, Regulations and Decisions, retrieved from the Official Government Gazette, where Greek legislation is published. This collection is one of the official, publicly available sources of classified Greek legislation suitable for our classification task.\footnote{Another official source is that of European Legislation written in modern Greek, available at: \href{https://eur-lex.europa.eu/browse/directories/legislation.html}{https://eur-lex.europa.eu/browse/directories/legislation.html}.}

Currently, the original catalogue is publicly offered in MS Word (.doc) format through the portal e-Themis\footnote{The portal is hosted at \href{https://www.secdigital.gov.gr/e-themis/}{https://www.secdigital.gov.gr/e-themis/}, where you can find the thematic index (hierarchy).}, the legal database and management service of it, under the administration of the Ministry of the Interior (Affairs)\footnote{\href{https://www.ypes.gr}{https://www.ypes.gr}}. E-Themis is primarily focused on providing legislation on a multitude of predefined thematic categories, as described in the catalogue. The main goal is to help users find legislation of interest using the thematic index.

\begin{figure}[t]
    \centering
    \includegraphics[width=\linewidth]{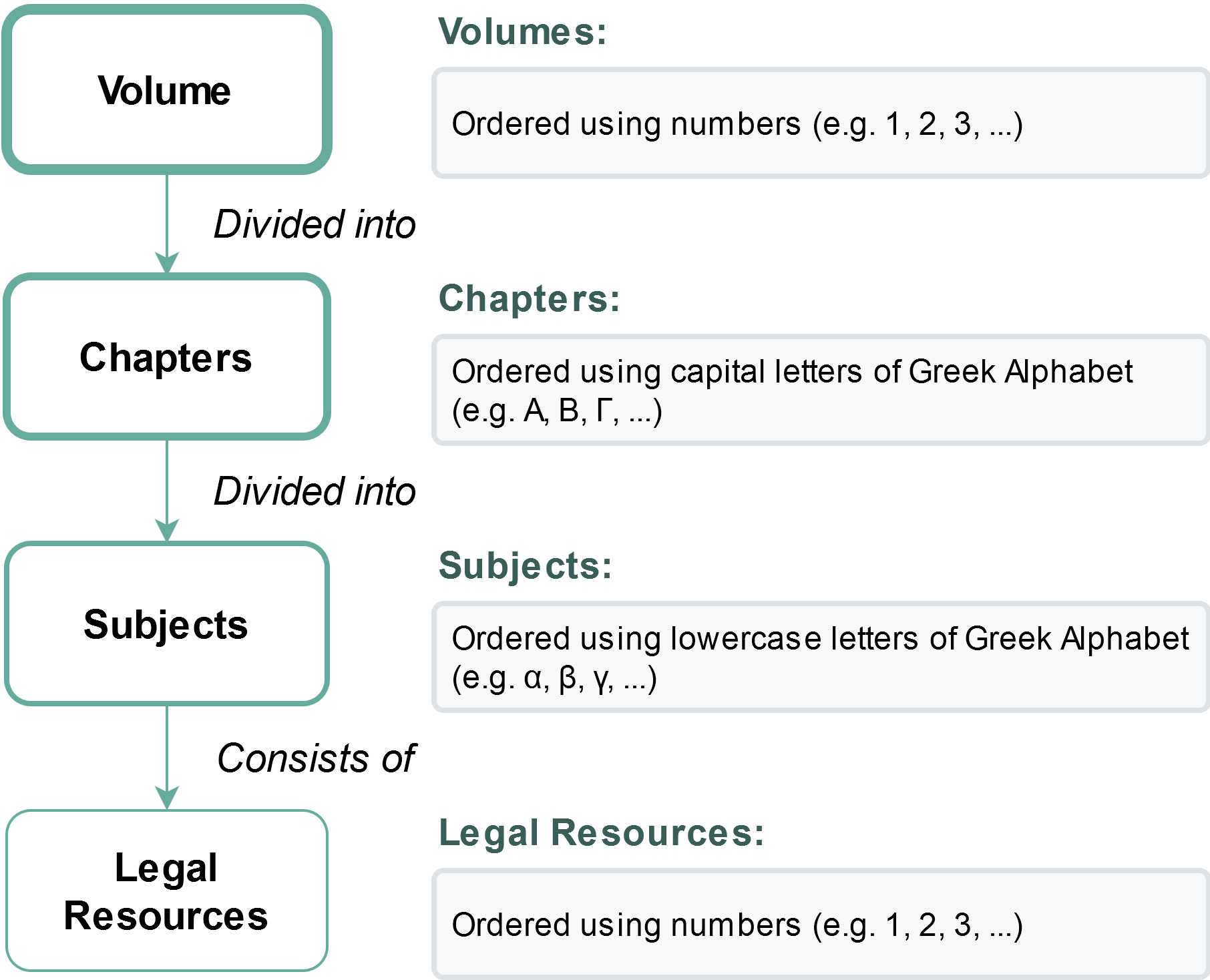}
    \captionsetup{justification=centering}
    \caption[GLC thematic hierarchy]{Original GLC thematic hierarchy}
    \label{fig:rap_hier}
    \vspace{-4mm}
\end{figure}

The original collection follows a bibliographic structure (\autoref{fig:rap_hier}).\footnote{In fact, it originated through a proper printed thesaurus.} It consists of 47 legislative \textit{volumes} and each volume corresponds to a main thematic topic. Inside each volume, the main thematic topic is divided into thematic subcategories which are called \textit{chapters} and subsequently, each chapter breaks down to \textit{subjects} which contain the legal resources. The total number of chapters is 389 while the total number of subjects is 2285, creating an interlinked thematic hierarchy. Thus the task is defined as a \emph{multi-level text classification task}, where the goal is to predict the thematic category at each level (volume, chapter, subject).

An example of this thematic hierarchy is the volume of ``Criminal Law'', which is divided into 9 chapters. ``International Criminal Law'' and ``Military Criminal Law'' are two out of the nine chapters. Subsequently, ``International Criminal Law'' is subdivided into 8 subjects (e.g., ``Genocide'', ``Counterfeiting'' etc.) and ``Military Criminal Law'' is subdivided into 4 subjects (e.g., ``Military Criminal Code'', ``Legal Remedies'' etc.). 
Another example is the volume of ``Labour Law'', divided into 17 chapters. Two of those are ``Collective Employment Contracts'' and ``Hygiene and Safety of Employees''. ``Collective Employment Contracts'' is subdivided into 7 subjects (e.g., ``Limits of Salaries'', ``Holiday Allowance'' etc.) and ``Hygiene and Safety of Employees'' is subdivided into 3 subjects (e.g., ``Hygiene And Safety Of Workplaces And Employees'', ``Work Health Books'' etc.).

For the study presented in this paper, we curated and publicly release a new dataset named Greek Legal Code (\textit{GLC}), containing all the available legislation from the original catalogue in JSON format. Each JSON file contains a legal resource (law) along with its metadata, i.e., thematic topics, publication year, document type, etc., as they were extracted from the original documents.\footnote{In Appendix~\ref{sec:appendix_a}, we present in great detail the pre-processing procedure.}

\subsection{Dataset Statistics}

In this section, we present a detailed quantitative analysis of the final dataset, consisting of 47,563 documents (i.e., categorized legal resources). 
\paragraph{Data Split} \textit{GLC} is split into three subsets: training (60\%), development (20\%) and test (20\%), as shown in \autoref{stats:tabl1}. The documents are distributed uniformly for all levels of the class hierarchy in order to achieve the same level of partitioning from bottom to top (i.e., from each subject to the whole dataset). 

\begin{table}[H]
\centering
\resizebox{\columnwidth}{!}{
\begin{tabular}{|c|c|c|c|}
\hline
\textbf{Subset} & \textbf{Docs} & \textbf{Mean \# of tokens / doc} & \textbf{(\textless{}100 tokens)} \\ \hline
Train  (60\%)   & 28536                  & 600                      & 15412 (54\%)                                   \\ \hline
Dev.   (20\%)   & 9511                   & 574                      & 5175 (54.4\%)                                  \\ \hline
Test.   (20\%)  & 9516                   & 595                      & 5075 (53.3\%)                                  \\ \hline
\textbf{Total:} & 47563                  & 594                      & 25662 (54\%)                                   \\ \hline
\end{tabular}
}
\caption{Dataset split and statistics.}
\label{stats:tabl1}
    \vspace{-6mm}
\end{table}

\begin{figure}[h]
    \centering
    \includegraphics[width=\linewidth]{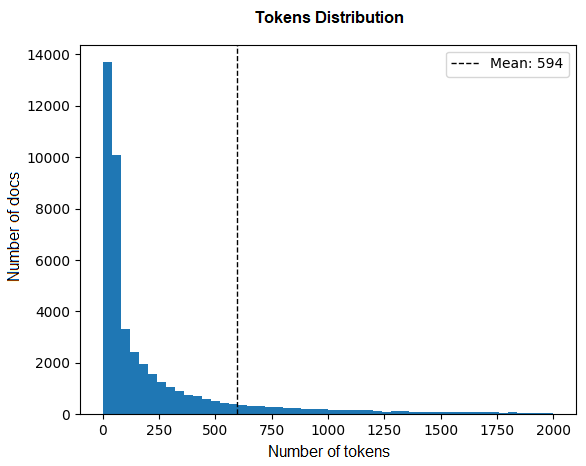}
    \captionsetup{justification=centering}
    \caption[GLC tokens' distribution]{Tokens' distribution over docs in GLC}
    \label{stats:meantokens}
    \vspace{-4mm}
\end{figure}

\paragraph{Document Size}  In Figure~\ref{stats:meantokens}, we observe that more than half of the documents in our dataset have less than 100 tokens. We see that most of the documents are not that long, with their total number of tokens being below the mean number of tokens across all documents. Many records (documents) in GLC, especially the older ones have only the descriptive title of the legal resource or a small part of it and not the full text.

\begin{table}[H]
\centering
\resizebox{\columnwidth}{!}{
\begin{tabular}{|c|c|c|c|c|}
\hline
\textit{} & \textbf{Total} & \textbf{Frequent} & \textbf{Few-shot (\textless{}10)} &\textbf{Zero-shot} \\ \hline
\textbf{Volume}            & 47             & 47 (100\%)        & 0                                 & 0             \\ \hline
\textbf{Chapter}           & 389            & 333 (85.6\%)      & 53 (13.6\%)                       & 3 (00.7\%)    \\ \hline
\textbf{Subject}           & 2285           & 712 (31.2 \%)     & 1431 (62.6\%)                     & 142 (06.2\%)  \\ \hline
\end{tabular}}
\caption{Number of classes per thematic level and their distribution to frequent-few-zero categories}
\label{stats:tabl2}
\vspace{-4mm}
\end{table}

\paragraph{Label Frequency}    
\textit{GLC} classes are divided into three categories for each thematic level: \textit{frequent classes}, which occur in more than 10 training documents and can be found in all three subsets (training,  development and test); \textit{few-shot classes}\footnote{The terms few-shot and zero-shot are used to express the under-representation of classes in \textit{GLC} rather than transfer learning approaches as in other deep learning literature.} which appear in 1 to 10 training documents and also appear in the documents of the development and test sets, and \textit{zero-shot classes} which appear in the development and/or test, but not in the training documents. As demonstrated in \autoref{stats:tabl2}, many classes are under-represented, especially in the thematic level of subjects, causing the appearance of few- and zero-shot categories. The appearance of underrepresented classes increases as we move towards more specific thematic levels. \autoref{stats:tabl3} shows the total number of documents per category and thematic level.

\begin{table}[H]
\centering
\resizebox{\columnwidth}{!}{
\begin{tabular}{|c|c|c|c|c|}
\hline
\textit{}        & \textbf{Total} & \textbf{Frequent} & \textbf{Few-shot (\textless{}10)} &\textbf{Zero-shot} \\ \hline
\textbf{Volume}  & 47563                & 47563 (100\%)     & 0                                 & 0                      \\ \hline
\textbf{Chapter} & 47563                & 47108 (99.0\%)    & 445 (00.9\%)                      & 10 (\textless{}00.1\%) \\ \hline
\textbf{Subject} & 47563                & 38475 (80.9\%)    & 8870 (18.6\%)                     & 218 (00.5\%)           \\ \hline
\end{tabular}
}
\caption{Number of documents as labeled per thematic level and their distribution to frequent-few-zero categories}
\label{stats:tabl3}
\end{table}

In the volume level, all the classes belong to the frequent category and are sufficiently represented, as more than 10 documents per class exist in the training data. 
In the chapter level, few-shot classes appear and are rather underrepresented as most documents are classified among frequent classes, leaving less than 1\% of the total documents to be associated with \textasciitilde14\% of the total classes.
In the subject level, data are even more unequally distributed over classes. The majority of documents are classified into frequent classes, leaving more than half of the total classes (\textasciitilde63\%) to be associated with less than 20\% of the total documents, along with 142 classes having zero representation at the training subset.

\begin{figure*}[t]
    \centering
    \includegraphics[width=\textwidth]{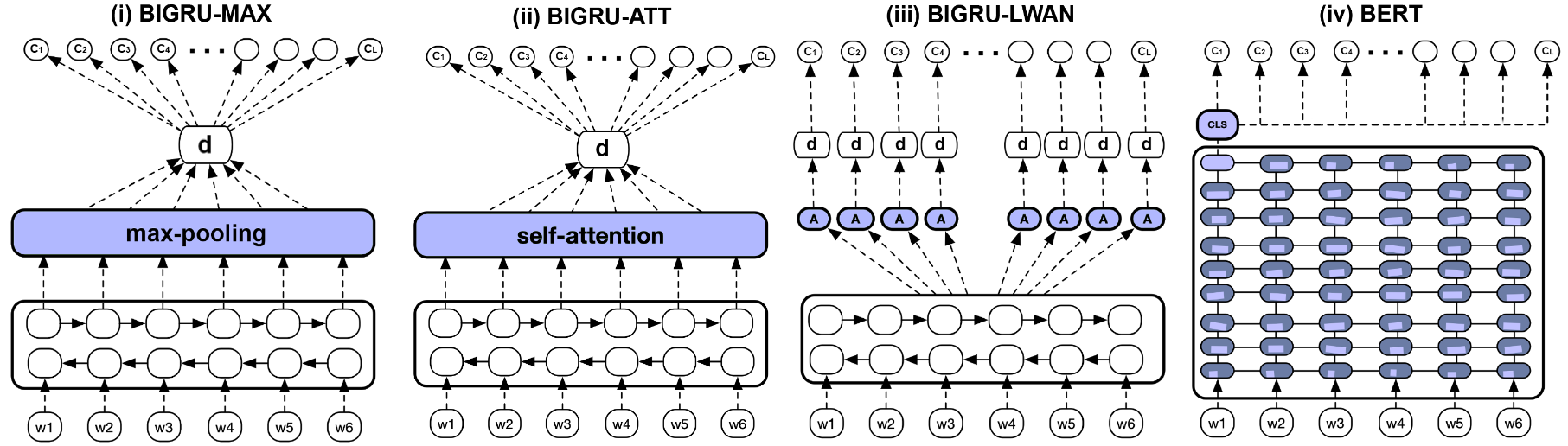}
    \captionsetup{justification=centering}
    \caption[Illustration of architectures]{Illustration of (i) BIGRU-MAX, (ii) BIGRU-ATT, (iii) BIGRU-LWAN and (iv) BERT}
    \vspace{-3mm}
\end{figure*}

\section{Task and Methods}\label{sec:task}

\subsection{Multi-class Legal Topic Classification for Greek Legislation}

We study multi-class legal topic classification for Greek legislation using the 
\textit{GLC} dataset\footnote{Available at: \href{https://huggingface.co/datasets/greek\_legal\_code}{https://huggingface.co/datasets/greek\_legal\_code}}. 
\textit{GLC} offers 3 hierarchical levels of thematic categorization, forming a tree structure that follows the original GLC organization as explained in Section 3.1. We divide the classification task for \textit{GLC} into three separate classification tasks. Each task deals with a different level of the thematic hierarchy (i.e., volume, chapter or subject) and the classifier's objective is to predict the correct class out of all the classes in this specific thematic level. 

\subsection{Methods}

In our study, we consider an arsenal of nine methods ranging from traditional machine learning techniques (2) and RNN-based methods (3), to state-of-the-art Transformer-based methods (4).\vspace{2mm}

\noindent\textbf{SVM-BOW:} Support Vector Machines (SVMs) is a strong baseline for the multi-class classification problem as it is one of the highest performing traditional ML methods. We represent the legal documents using Bag-of-Words (BoW) features, getting the most frequent n-grams across all training data, weighted by TF-IDF.\vspace{2mm}

\noindent\textbf{XGBOOST-BOW:} XGBoost ~\cite{Chen2016XGBoostAS} is a scalable, fast and robust open-source implementation\footnote{Available at: \href{https://github.com/dmlc/xgboost/}{https://github.com/dmlc/xgboost/}} of the Gradient Boosting decision tree algorithm. In XGBoost, the model is fitted on the gradient of loss generated from the previous step and the gradient boosting algorithm is modified so that it works with any differentiable loss function. In our case, the documents are represented using BoW features weighted by TF-IDF (similarly to SVM-BOW).\vspace{2mm} 

\noindent\textbf{BIGRU-MAX:} The first RNN-based method that we examine is BiGRU with max-pooling, employing pre-trained, domain-specific word embeddings ~\cite{Angelidis2018NamedER}.\footnote{\citet{Angelidis2018NamedER} released a Word2Vec model trained on Greek legal corpora, including Greek national and EU legislation.} The stacked BiGRU encoder converts the pre-trained word embeddings \textbf{w$_i$} into context-aware embeddings \textbf{h$_i$}.
Context-aware token embeddings pass through a max-pooling layer to produce the final document representation \textbf{d} by reducing the initial matrix (token-wise vectors) into a single vector but hopefully, keep the most salient information.
Finally, a dense layer with \textit{L} output units and the \textit{softmax()} activation is deployed, to transform the document representation \textbf{d} into a probability distribution over \textit{L} classes. \textit{L} is defined according to the examined task, i.e., \textit{L}=47 for the volume-level task, \textit{L}=389 for the chapter-level task and \textit{L}=2285 for the subject-level task.\vspace{2mm}

\noindent\textbf{BIGRU-ATT:} The second RNN-based method is a BiGRU network with self-attention ~\cite{Xu2015ShowAA, Chalkidis2019LargeScaleMT}. This methods uses the very same word embedding and BiGRU encoder, similarly to the previously described method (BIGRU-MAX). Instead of max-pooling, it uses the self-attention mechanism to produce a final document representation \textbf{d}. This representation is computed as the weighted sum of the BiGRU context-aware embeddings \textbf{h$_i$}, weighted by the self-attention scores \textit{a$_i$} produced as:

\begin{equation}
a_{i}=\frac{\exp \left(\mathbf{h}_{i}^{\top} \mathbf{u}\right)}{\sum_{j} \exp \left(\mathbf{h}_{j}^{\top} \mathbf{u}\right)}
\end{equation}

\begin{equation}
\mathbf{d}=\frac{1}{T} \sum_{i=1}^{T} a_{i} \mathbf{h}_{i}
\end{equation}

\textit{T} represents the document’s length in words while \textbf{u} is a trainable vector used to compute the attention scores \textit{a$_i$} over \textbf{h$_i$}.
Similarly to BIGRU-MAX, a final dense layer with \textit{L} output units and \textit{softmax()} activation is deployed to predict the correct output class using a probability distribution over all the classes.\vspace{2mm}

\noindent\textbf{BIGRU-LWAN:} The third RNN-based method replaces the self-attention mechanism of BIGRU-ATT with the label-wise attention mechanism. The original Label-Wise Attention Network (LWAN) was introduced in ~\citet{Mullenbach2018ExplainablePO} using a CNN-based encoder. Later on, ~\cite{Chalkidis2019LargeScaleMT,Chalkidis2019ExtremeML} replaced it with a BiGRU encoder.
In contrast with BIGRU-ATT, this label-wise attention technique uses \textit{L} independent attention heads, one per class, generating \textit{L} label-wise document representations \textbf{d$_l$} from the sequence of \textbf{h$_i$} vectors produced by the BiGRU encoder. The intuition is that each label-wise document embedding is dedicated in predicting the corresponding class, focusing on possibly different aspects of each representation \textbf{h$_i$}. In effect, different parts of the representation may be more relevant for different classes.

\begin{equation}
a_{l i}=\frac{\exp \left(\mathbf{h}_{i}^{\top} \mathbf{u}_{l}\right)}{\sum_{i^{\prime}} \exp \left(\mathbf{h}_{i^{\prime}}^{\top} \mathbf{u}_{l}\right)}
\end{equation}

\begin{equation}
\mathbf{d}_{l}=\frac{1}{T} \sum_{i=1}^{T} a_{l i} \mathbf{h}_{i}
\end{equation}

Again, \textit{T} represents the document’s length in words while \textbf{u$_l$} ($l=1, 2, ..., L$) is a trainable vector used to compute the attention scores \textit{a$_{l i}$} of the \textit{l}-th attention head. Then, each attention head \textbf{d$_l$} goes through an independent dense layer for each class and similarly to the previous methods, \textit{softmax()} activation is deployed to transform the document representation \textbf{d} into a probability distribution over \textit{L} classes.\vspace{2mm}

\noindent\textbf{M-BERT:} BERT ~\cite{Devlin2019BERTPO} is a Transformer-based ~\cite{Vaswani2017AttentionIA} language model initially developed by Google. In BERT, deep bidirectional representations are pre-trained from unlabeled text by jointly conditioning on both left and right context. As a result, for any new task, the pre-trained BERT model can be fine-tuned with just one additional task-specific output layer trained with task-specific data. We employ the multilingual version of the 12-layer BERT-BASE-CASED model, which supports modern Greek, alongside 99 languages, out-of-the-box. We add linear layer on top of M-BERT encoder followed by a \textit{softmax()} activation. This extra dense layer is fed with the so-called ``classification token'' ([cls]) of the BERT encoder as described in ~\cite{Devlin2019BERTPO}, serving as the final document representation.\vspace{2mm}

\noindent\textbf{XLM-ROBERTA:} The next BERT-based model we examine is XLM-RoBERTa ~\cite{Conneau2020UnsupervisedCR}, a multilingual adaptation of RoBERTa ~\cite{Liu2019RoBERTaAR}. RoBERTa is built following BERT’s architecture, while removed the next-sentence pre-training objective and trained with much larger batches and higher learning rate. Based on this study, ~\citet{Conneau2020UnsupervisedCR} proposed the XLM-RoBERTa model, which supports 100 different languages (Greek included) and is trained on 2.5TB of filtered Common Crawl data.\vspace{2mm}

\noindent\textbf{GREEK-BERT:} We also experiment with GREEK-BERT ~\cite{Koutsikakis2020GREEKBERTTG}, a native monolingual version of BERT, trained solely on modern Greek, achieving state-of-the-art results in several NLP tasks. GREEK-BERT was pre-trained on 29GB of text from a corpus consisting of the Greek part of Wikipedia, the Greek part of the European Parliament Proceedings Parallel Corpus (Europarl) \cite{Koehn2005EuroparlAP} and OSCAR \cite{OrtizSuarezSagotRomary2019}, a clean version of Common Crawl. Even though multilingual models like the previous two offer exceptional performance, monolingual models usually surpass them in most downstream tasks. Again, GREEK-BERT follows the previous BERT models configuration.\vspace{2mm}

\noindent\textbf{GREEK-LEGAL-BERT:} Finally, we experiment with GREEK-LEGAL-BERT~\cite{athinaios2020}, another BERT-flavored model for modern Greek. Its pre-training corpus is based entirely on Greek legal documents and consists of around 5GB of Greek and EU legislation documents retrieved by Nomothesia\footnote{\href{http://legislation.di.uoa.gr}{http://legislation.di.uoa.gr}.} \cite{Chalkidis2017ModelingAQ}, a Greek legislative Knowledge Base. It also follows the previously described configuration.

\begin{table}[th]
\centering
\resizebox{\linewidth}{!}{
\begin{tabular}{|l|l|}
\hline
\textbf{Method} & \textbf{Hyperparameters}                                                                                                                                                                                \\ \hline
SVM-BOW         & \begin{tabular}[c]{@{}l@{}}Kernel: {[}linear, rbf{]}\\ C: {[}0.25, 0.50, 0.75, 1{]}\\ N-Grams: {[}(1,3), (1,5){]}\\ Max-Features: {[}200k, 400k{]}\end{tabular}                                          \\ \hline
XGBOOST-BOW     & \begin{tabular}[c]{@{}l@{}}Booster: gbtree with softmax\\ N-Estimators: 800\\ Max-Depth: {[}4, 5, 7, 10{]}\\ Min-Child-Weight: {[}2, 5, 10{]}\end{tabular}                                               \\ \hline
BIGRU-*         & \begin{tabular}[c]{@{}l@{}}HYPEROPT\footnotemark on:\\ Layers: {[}1, 2{]}\\ Units: {[}200, 300, 400{]}\\ Dropout: {[}0.1, 0.2, 0.3, 0.4{]}\\ Word Dropout: {[}0, 0.01, 0.02{]}\\ Batch Size: {[}8, 16{]}\end{tabular} \\ \hline
*BERT*          & \begin{tabular}[c]{@{}l@{}}Dropout: 0.1\\ Learning Rate: {[}1e-5, 2e-5, 3e-5, 5e-5{]}\\ Batch Size: 8\end{tabular}                                                                                       \\ \hline
\end{tabular}
}
\vspace{-2mm}
\caption{Hyper parameters search space for all models}
\label{exp:hyperparams}
\vspace{-6mm}
\end{table}
\footnotetext{See: \href{https://hyperopt.github.io/hyperopt/}{https://hyperopt.github.io/hyperopt/}}

\begin{table*}[t]
\centering
\resizebox{\textwidth}{!}{
\begin{tabular}{l|cccccc|ccccccccc}
\hline
\hline
\multicolumn{1}{c}{\textsc{MODEL}} & \multicolumn{6}{c}{\emph{Volume (47 classes)}} & \multicolumn{9}{c}{\emph{Chapter (389 classes)}} \\
    \hline
    & \multicolumn{3}{c}{\textsc{ALL LABELS}} & \multicolumn{3}{c}{\textsc{FREQUENT}}
    & \multicolumn{3}{c}{\textsc{ALL LABELS}} & \multicolumn{3}{c}{\textsc{FREQUENT}} & \multicolumn{3}{c}{\textsc{FEW-SHOT}} \\
    \hline
    & P & R & F1 & P & R & F1 & P & R & F1 & P & R & F1 & P & R & F1 \\
    \cline{2-16}                            
    \textsc{SVM-BOW}                & 85.3 & 85.3 & 85.3 & 85.3 & 85.3 & 85.3  & 77.9 & 77.9 & 77.9 & 77.9 & 78.6 & 78.2 & \textbf{90.0} & 09.3 & 16.8 \\
    \textsc{XGBOOST-BOW}            & 77.2 & 77.2 & 77.2 & 77.2 & 77.2 & 77.2  & 67.5 & 67.5 & 67.5 & 67.8 & 68.1 & 67.9 & 19.2 & 10.3 & 13.4 \\
    \hline
    \textsc{BIGRU-MAX}              & 84.3 & 84.3 & 84.3 & 84.3 & 84.3 & 84.3  & 77.5 & 77.5 & 77.5 & 77.9 & 77.9 & 77.9 & 44.9 & 45.4 & 45.1 \\
    \textsc{BIGRU-ATT}              & 86.4 & 86.4 & 86.4 & 86.4 & 86.4 & 86.4  & 81.1 & 81.1 & 81.1 & 81.1 & 81.6 & 81.3 & 86.7 & 40.2 & 54.9 \\
    \textsc{BIGRU-LWAN}             & 84.1 & 84.1 & 84.1 & 84.1 & 84.1 & 84.1  & 76.8 & 76.8 & 76.8 & 76.9 & 77.3 & 77.1 & 63.8 & 30.9 & 41.7 \\
    \hline
    \textsc{M-BERT}           & 85.8 & 85.8 & 85.8 & 85.8 & 85.8 & 85.8  & 80.2 & 80.2 & 80.2 & 80.1 & 80.6 & 80.4 & 81.0 & 43.9 & 56.6 \\
    \textsc{XLM-ROBERTA}            & 85.0 & 85.0 & 85.0 & 85.0 & 85.0 & 85.0 & 80.6 & 80.6 & 80.6 & 80.6 & 81.0 & 80.8 & 80.3 & 37.9 & 51.3 \\
    \textsc{GREEK-BERT}             & 87.5 & 87.5 & 87.5 & 87.5 & 87.5 & 87.5 & 82.4 & 82.4 & 82.4 & 82.4 & 82.8 & 82.6 & 81.0 & 45.2 & \textbf{57.8} \\
    \textsc{GREEK-LEGAL-BERT}       & \textbf{89.4} & \textbf{89.4} & \textbf{89.4} & \textbf{89.4} & \textbf{89.4} & \textbf{89.4} & \textbf{84.3} & \textbf{84.3} & \textbf{84.3} & \textbf{84.4} & \textbf{84.8} & \textbf{84.6} & 79.0 & \textbf{45.8} & \textbf{57.8} 
    \end{tabular}
}
\caption{Experimental results for \emph{Volume} and \emph{Chapter} levels. }
\vspace{-2mm}
\label{exp:exptbl}
\vspace{-3mm}
\end{table*}

\section{Experiments}

\subsection{Experimental SetUp}

We tune all methods by grid-searching over the core hyper-parameters (per method) presented in Table~\ref{exp:hyperparams}, except the RNN-based methods where we used HyperOpt due to the large search space. We select the models with the best performance on the development set. We use early stopping on the development loss using the Adam \cite{Kingma2015AdamAM} optimizer. All BERT models follow the base configuration, i.e., 12 layers with 786 units and 12 attention heads each.\footnote{All models are available at \href{https://huggingface.co/models}{https://huggingface.co/models}.} We report micro-averaged Precision (P), Recall (R), and F1-score (F1) on the test set. For each method, we perform 5 runs with different seeds and report the average. We report results for the frequent and few-shot classes (when available), omitting results for zero-shot classes since our methods are incapable of zero-shot learning.
We release our code for reproducibility:\footnote{See: \href{https://github.com/christospi/glc-nllp-21}{https://github.com/christospi/glc-nllp-21}.}

\subsection{Volume-level Classification Evaluation}

In volume-level classification, all the 47 possible classes are sufficiently represented and belong to the frequent category, while the mean number of documents per class is approximately 1k. However, we acknowledge class imbalance (label skewness) as a challenge in our dataset, though not so evident here. 
Table \ref{exp:exptbl} (first zone) demonstrates the results for this task.
SVM-BOW proves to be a strong competitor in the first classification task (85.3 in F1).
Interestingly, it outperforms two of our neural methods with domain-specific word embeddings, namely BIGRU-MAX and BIGRU-LWAN with 84.3 and 84.1 F1 scores, respectively. As for XGBoost, although it seemed quite promising and very fast at training, its inadequate F1 score of 77.2 places it at the bottom of the list for this task. 

Among the RNN-based neural methods, BIGRU-ATT achieves an F1 score of 86.4 outperforming the rest of the RNN-based methods. 
Its results indicate the significance of two of its fundamental features: (i) the domain-specific word2vec embeddings and (ii) the cumulative self-attention head that provides an advantageous final document representation. 
Compared to BIGRU-MAX, we believe that its max-pooling layer hinders some of the document’s particularities and thus, it yields a lower score. Likewise, the BIGRU-LWAN method with L different attention heads seems to be more tailor-made for multi-label classification tasks, as it does not offer any performance improvement compared to BIGRU-ATT.

GREEK-LEGAL-BERT proves to be the best method we experimented with, achieving a score of 89.4 in F1, followed by the generic GREEK-BERT with a score of 87.5 in F1. The two multilingual models (M-BERT and XLM-ROBERTA) also achieve satisfying F1 scores (85.8 and 85.0), confirming their claim to offer top-notch results in most downstream NLP tasks.
The results demonstrate that monolingual models are able to surpass other advanced multilingual transformer-based models. We hypothesize that the fact that GREEK-LEGAL-BERT is pre-trained entirely on Greek legal corpora is the main reason of its superiority, while the specialization in the Greek language, alongside GREEK-BERT, seems even more critical.

\subsection{Chapter-level Classification Evaluation}

In Table \ref{exp:exptbl} (second zone) we observe again that XGBOOST-BOW has the lowest performance with 67.5 in overall F1 score with SVM-BOW being the second worst overall (77.9) alongside BIGRU-MAX. The ranking of BERT-based models is similar to the previous task (Volume-level classification), i.e., monolingual models outperform their multilingual counterparts, while GREEK-LEGAL-BERT is better by 2\% compared to the generic GREEK-BERT model. These results further supporting our intuition on the importance of native language support and domain knowledge.

Inspecting the few-shot label results, we observe that BIGRU-ATT and BERT models outperform BIGRU-MAX by 10\%, while the traditional machine learning approaches have terrible performance (approx. 15 F1). These results highlight the importance of the attention mechanism and its capability to focus in specific parts of the text that are more prominent in relation to the downstream task and the specific labels.

\subsection{Subject-level Classification Evaluation}

In the last sub-task, where there are many more classes (2285 in total), we observe a similar ranking in the methods (Table~\ref{exp:exptbl2}). However, the absolute difference in few-shot labels' scores is higher this time, when comparing either the attention-based method with BIGRU-MAX (approx. +20\% F1) or monolingual with multilingual BERT models (approx. +9\% F1).

\begin{table}[t]
\centering
\resizebox{\columnwidth}{!}{
\begin{tabular}{l|ccccccccc}
\hline
\hline
\multicolumn{1}{c}{\textsc{MODEL}} & \multicolumn{9}{c}{\emph{Subject (2285 classes)}} \\
    \hline
    & \multicolumn{3}{c}{\textsc{ALL LABELS}} & \multicolumn{3}{c}{\textsc{FREQUENT}} & \multicolumn{3}{c}{\textsc{FEW-SHOT}} \\
    \hline
    & P & R & F1 & P & R & F1 & P & R & F1 \\
    \cline{2-10}                            
    \textsc{SVM-BOW}                & 37.9 & 37.9 & 37.9 & 37.9 & 47.8 & 42.3 & 00.0 & 00.0 & 00.0 \\
    \textsc{XGBOOST-BOW}            & 55.3 & 55.3 & 55.3 & 56.1 & 64.8 & 60.1 & 46.9 & 19.1 & 27.2 \\
    \hline
    \textsc{BIGRU-MAX}              & 62.9 & 62.9 & 62.9 & 66.0 & 70.5 & 68.1 & 47.1 & 37.8 & 42.0 \\
    \textsc{BIGRU-ATT}              & 74.8 & 74.8 & 74.8 & 75.3 & 79.6 & 77.4 & 72.6 & 61.1 & 66.3 \\
    \textsc{BIGRU-LWAN}             & 65.2 & 65.2 & 65.2 & 68.1 & 72.8 & 70.4 & 50.7 & 40.4 & 45.0 \\
    \hline
    \textsc{M-BERT}                 & 76.8 & 76.8 & 76.8 & 79.8 & 82.5 & 81.1 & 64.4 & 59.7 & 62.0 \\
    \textsc{XLM-ROBERTA}            & 78.0 & 78.0 & 78.0 & 80.3 & 83.4 & 81.8 & 68.2 & 62.1 & 65.0 \\
    \textsc{GREEK-BERT}             & 79.4 & 79.4 & 79.4 & 80.5 & 83.9 & 82.2 & \textbf{74.6} & 67.0 & 70.6 \\
    \textsc{GREEK-LEGAL-BERT}       & \textbf{81.2} & \textbf{81.2} & \textbf{81.2} & \textbf{83.0} & \textbf{85.5} & \textbf{84.2} & 73.6 & \textbf{69.4} & \textbf{71.4} \\
    \end{tabular}
}
\vspace{-2mm}
\caption{Experimental results for \emph{Subject} level. }
\label{exp:exptbl2}
\vspace{-3mm}
\end{table}

\subsection{General Observations}

There are two main observations:
\vspace{-2mm}
\begin{itemize}[leftmargin=8pt, itemsep=1pt]
    \item Pre-trained Transformer-based models perform exceptionally and the performance gains compared to traditional machine learning methods and RNN-based methods are increasing in relation to the number of labels. We can only speculate the following: (a) sub-word units used by Transformer-based models potentially play an important role in morphological rich languages like Greek, compared to full words processed by the rest of the examined models; and (b) the multi-head attention mechanism can better distill important detailed (specialized) information to distinguish categories (labels).
    \item The Label-wise Attention Network (LWAN) under-performs in the newly introduced multi-class text classification task. Again, we can only speculate that the produced label-wise document representations that improve results over the standard universal attention mechanism in the multi-label setting \cite{chalkidis-etal-2020-empirical} lead to a ``greedy'' over-scoring (produced logits) followed by an aggressive label competition (softmax over logits) in the examined multi-task setting leading to negative (poor) results. 
\end{itemize}

We aim to further review and validate these observations, while also study the reasons behind those thoroughly in the future, possibly using more datasets and ablation studies (e.g., train RNN-based methods with pretrained sub-word embeddings).

\section{Conclusions and Future Work}
\label{sec:conc}

We introduced Greek Legal Code (\textit{GLC}), a new publicly available dataset consisting of 47k Greek legislation resources. Relying on this dataset, we experimented with several classifiers, ranging from traditional machine learning and recurrent models to state-of-the-art transfer learning models. Through their performance evaluation, we realized that although traditional machine learning classifiers (e.g., SVM-BOW) set strong baselines for some of the considered tasks, they fall short against more sophisticated methods. In contrast, RNN-based methods relying on BiGRUs provide improved overall performance and were competitive to multilingual transformer-based architectures (M-BERT, XLM-ROBERTA).
Beyond doubt, monolingual transformer-based models (GREEK-BERT and GREEK-LEGAL-BERT) lead to state-of-the-art results, especially when they are pre-trained on in-domain corpora.

Nonetheless, more emphasis should be given to the qualitative and quantitative characteristics of the examined datasets. Intricacies like class imbalance, data scarcity and diversity apparently need special handling. Regarding our study, we noticed that few-shot and especially zero-shot classes need to be properly handled with appropriate methods, as standard classifiers are insufficient. As for the recent trend to develop novel monolingual BERT-based models, results show that already established multilingual models are incredibly powerful even in monolingual tasks. While research is on-going and these models are continuously being improved, also taking into consideration the computational costs, it is quite challenging to motivate researchers into making an effort to train monolingual models for medium or small-sized languages; especially when multilingual models can perform equally well or occasionally, even better.

In future work, we plan to investigate specialized methods with improved few-shot and zero-shot performance~\cite{Hu2018FewShotCP,Rios2018FewShotAZ,Chalkidis2019LargeScaleMT} that leverage various data properties (e.g., label descriptors and label hierarchy). Also, we intend to apply deep learning techniques that take into account the hierarchy of classes in datasets like \textit{GLC}, where there is an underlying taxonomy ~\cite{Kowsari2017HDLTexHD,Chalkidis2020AnES,Manginas2020LayerwiseGT}. Furthermore, experimenting with similar datasets like that of EU Legislation written in Greek will allow us to confirm our current conclusions. For example in the future, we could consider Cypriot legislation\footnote{\href{http://www.cylaw.org/nomoi/}{http://www.cylaw.org/nomoi/}} to evaluate the out-of-domain generalization of models trained on Greek legislation. Finally, our long term goal is to support and encourage further research in NLP for the Greek language by publishing novel datasets, introducing and experimenting with state-of-the-art methods.

\section*{Acknowledgements}
This work has received funding from the European Union's Horizon 2020 research and innovation programme under grant agreement No 825258.
This work is also partly funded by the Innovation Fund Denmark (IFD)\footnote{\url{https://innovationsfonden.dk/en}} under File No.\ 0175-00011A.

\bibliography{anthology,custom}
\bibliographystyle{acl_natbib}
\appendix

\section{Dataset Curation}
\label{sec:appendix_a}

We describe in-detail the processing method we followed to generate \textit{GLC} and describe the data structure of the final JSON documents.

\subsection{Parsing the Original Documents}

The original legislative volumes are encoded in MS Word (.doc) format. While most of them follow the double-column format, there are cases where they also include text in single-column format or even include scanned documents or images as legal resources, making the initial data quite noisy. Considering our objective, these abnormalities should be revised, and all the additional metadata contained in the .doc file (e.g., font style, size, page margins) should be removed. Thus, converting these documents into plain text files was of significant importance. To achieve that, we used docx2txt\footnote{Available at: \href{https://github.com/ankushshah89/python-docx2txt/}{https://github.com/ankushshah89/python-docx2txt/}.}, a python utility that detects and extracts text from .doc files.

Examining the output text files, we encountered problematic samples that needed special handling. For example, we found significant keywords (e.g., “$\Theta$EMA” which means subject) missing from text or even having typos, subject IDs found inline with subject titles etc., mostly due to minor inaccuracies in the text conversion. Furthermore, multiple white spaces, multiple new lines and special or corrupted characters occurred in the text. To overcome these complications, we performed data cleaning using heuristics and regular expressions to produce neat text files that follow the same normalized structural pattern.

\begin{figure}[ht]
        \centering
        \includegraphics[width=\linewidth]{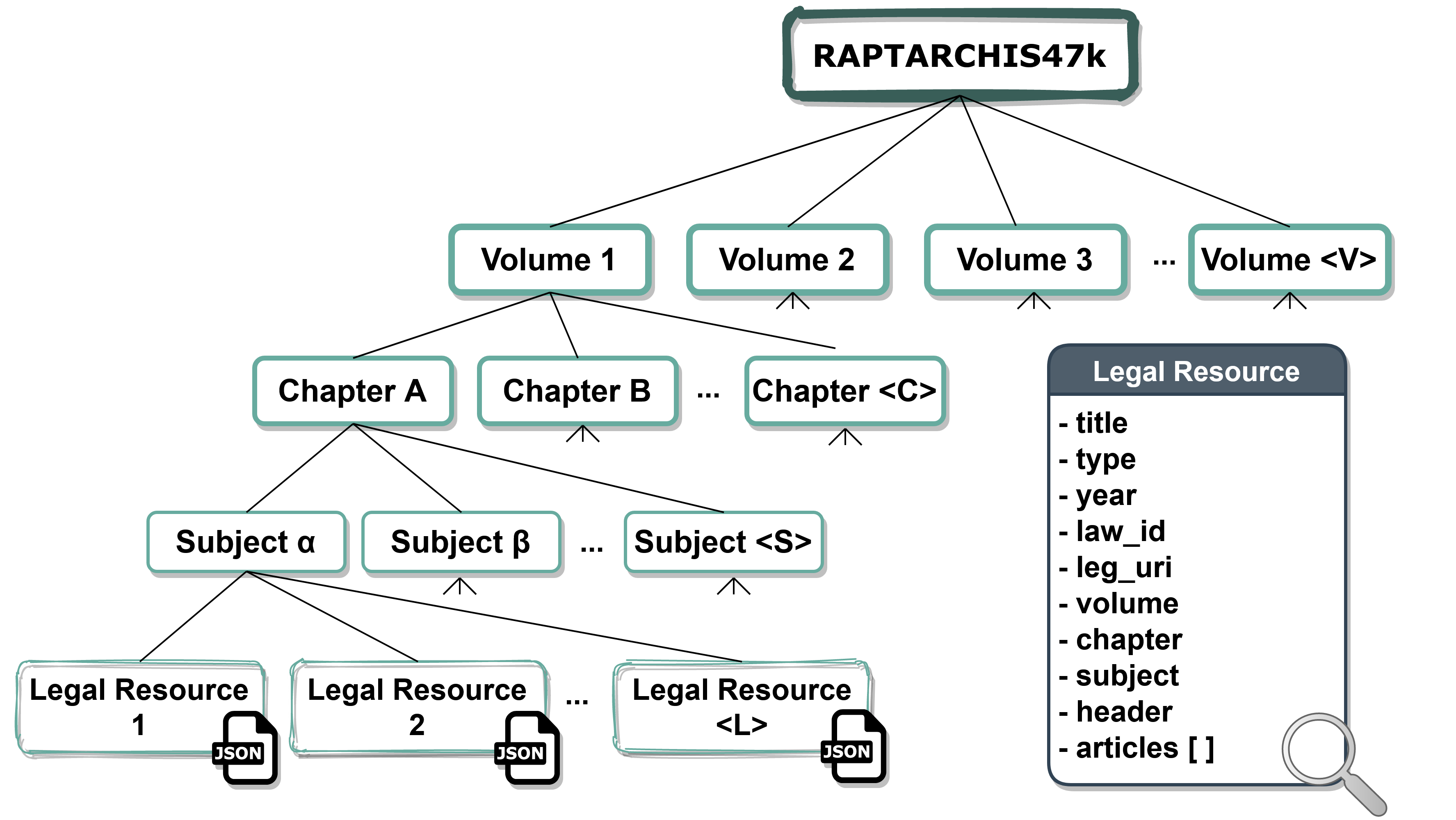}
        \captionsetup{justification=centering}
        \caption[Tree representation of GLC thematic hierarchy]{Tree representation of GLC thematic hierarchy}
        \label{fig:rap_tree}
    \end{figure}

Next, we built a rule-based parser in Python which receives these text files as input and produces JSON files that will be the inputs to our classifiers. The goal here is to separate legal resources into single documents, along with all their related metadata such as ID, publication year, title etc. and their classification hierarchy (i.e., in which volume, chapter and subject they belong to). Each final JSON file represents a unique legal resource, ready to be fed into the machine-learning models we built. The parser builds in memory a tree of depth 4 that represents the whole GLC hierarchy. The first level consists of the thematic volumes while the second level contains all the thematic chapters for each volume. The third level includes the thematic subjects of the individual chapters, and finally, the leaf nodes represent the legal resources. An overview of the tree can be found in \autoref{fig:rap_tree}.

\begin{figure}[ht]
        \centering
        \includegraphics[scale=0.50]{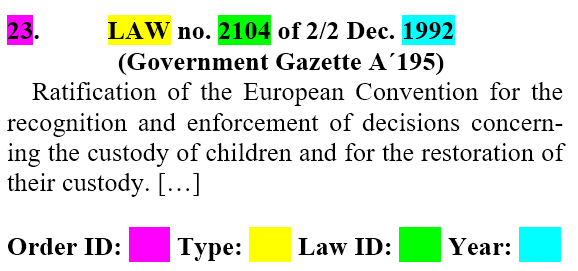}
        \captionsetup{justification=centering}
        \caption[Legal resource's metadata]{Legal resource's metadata (translated)}
        \label{fig:rap_meta}
    \end{figure}

With the use of regular expressions, the parser organizes the content to its thematic hierarchy. Each legal resource may contain the whole original legislative document, some articles of it or even a short sentence (usually its original title or a short description). Hence, the parser attempts to identify and separate any existing articles. However, if this is not feasible, it just keeps the whole body as a text chunk. No deeper parsing is performed (i.e., in paragraphs, sentences) as this is out of scope. The final step is to populate the leaf nodes (i.e., the legal resources) with the appropriate metadata and enhance the available text samples. To accomplish this, the parser uses the metadata fragment of each legal resource to extract the necessary information. Specifically, the words of interest are shown in \autoref{fig:rap_meta}, depicting an example of a metadata fragment.

Again, with proper regular expressions, the parser manages to retrieve the requested information. Also, having available the type, the year of publication and the ID of each legal resource, the parser is able to uniquely identify each one of them by using the following pattern: \textbf{\emph{\{type\}/\{year\}/\{id\}}}. Exploiting that, it searches for duplicate legal resources that may exist in the dataset. For example, one law may be present in more than one subject due to the thematic variety of its articles. To avoid any complexities and because our task is multi-class and not multi-label classification, the parser removes these resources entirely from the dataset.

Moreover, the parser manages to enhance the content of some legal resources (depending on their type\footnote{See the supported legislation types at: \href{http://legislation.di.uoa.gr/search/}{http://legislation.di.uoa.gr/search/}.}) by utilizing the platform Nomothesia\footnote{See: \href{http://legislation.di.uoa.gr/}{http://legislation.di.uoa.gr/}.}~\cite{Chalkidis2017ModelingAQ}. Nomothesia makes Greek legislation available as open linked data using semantic web technologies. Through its RESTful API and by adopting the following URI template:  \emph{http://www.legislation.di.uoa.gr/eli/\\\textbf{\{type\}/\{year\}/\{id\}}/data/json}
the parser manages to retrieve the text of any legal resource in JSON format, as offered through Nomothesia. Then, it compares the number of tokens of the original and the fetched text fragments and eventually keeps the more extensive. In that way, the parser succeeds in enhancing the size and quality of the dataset. 

\begin{figure}[H]
        \centering
        \includegraphics[scale=0.30]{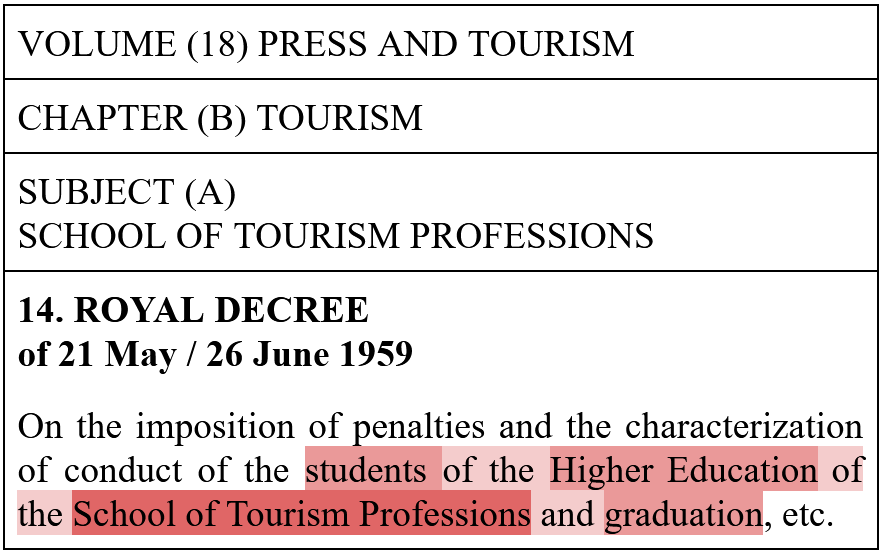}
        \captionsetup{justification=centering}
        \caption[Small-sized sample of GLC]{Small-sized sample of GLC indicating highly representative words (translated)}
        \label{fig:smallsample}
    \end{figure}

For each final document, the text content consists of the header along with the body. In case of successful parsing the body consists of multiple articles. Otherwise, it only contains a single text passage. Evaluating the final data, we noticed that many legal resources have limited tokens count (as shown in charts of Section 3.2). However, we consider this not to be a crucial problem since meaningful information (e.g., highly representative words) is quite dense in most of these samples as shown in \autoref{fig:smallsample}.

\begin{figure}[H]
        \centering
        \includegraphics[width=\linewidth]{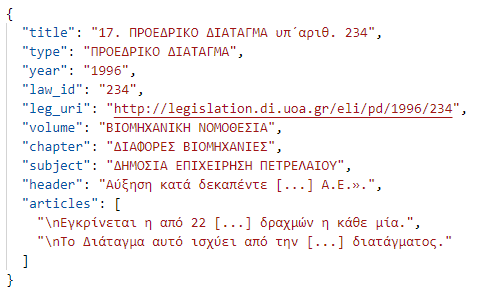}
        \captionsetup{justification=centering}
        \caption[Sample's transformation to JSON format]{Final legal resource as JSON. The legal resource has been parsed and enhanced with two articles, as fetched from Nomothesia web platform}
        \label{fig:json_sample}
    \end{figure}

Finally, the complete dataset, consisting of JSON files following the format of \autoref{fig:json_sample} is distributed to train, development and test subsets as described in Section 3.2.
    
\end{document}